\documentclass[a4paper]{article}
\usepackage{iclr2018_workshop,times}

\usepackage[english]{babel}
\usepackage[utf8x]{inputenc}
\usepackage[T1]{fontenc}

\usepackage[a4paper,top=3cm,bottom=2cm,left=3cm,right=3cm,marginparwidth=1.75cm]{geometry}

\usepackage[noblocks]{authblk}

\usepackage{amsmath}
\usepackage{graphicx}
\usepackage[colorinlistoftodos]{todonotes}
\usepackage{hyperref}
\usepackage{tabularx}
\usepackage{wrapfig}

\title{Natural Language Inference over Interaction Space: ICLR 2018 Reproducibility Report}
\author[1]{Martin Mirakyan}
\author[2]{Karen Hambardzumyan}
\author[3]{Hrant Khachatrian}
\affil[1,2,3]{YerevaNN}
\affil[1]{College of Science and Engineering, American University of Armenia}
\affil[2,3]{Department of Informatics and Applied Mathematics, Yerevan State University}

\begin{document}
\maketitle

\begin{abstract}
We have tried to reproduce the results of the paper ``Natural Language Inference over Interaction Space'' submitted to ICLR 2018 conference as part of the ICLR 2018 Reproducibility Challenge. Initially, we were not aware that the code was available, so we started to implement the networ from scratch. We have evaluated our version of the model on Stanford NLI dataset and reached 86.38\% accuracy on the test set, while the paper claims 88.0\% accuracy. The main difference, as we understand it, comes from the optimizers and the way model selection is performed.
\end{abstract}

\section{Densely Interactive Inference Networks}

The paper \citep{gong2018natural} describes a deep neural network that takes two sentences as inputs (\textit{hypothesis} and \textit{premise}) and outputs probabilities for three classes (entailment, contradiction and neutral). The network architecture is called Densely Interactive Inference Network (DIIN). We were not aware that the authors have released the codebase, so we decided to re-implement the architecture from scratch. The code for our implementation is available on GitHub\footnote{\url{https://github.com/YerevaNN/DIIN-in-Keras}}. We used Keras library \citep{chollet2015keras} with TensorFlow backend to implement and train the architecture.

The architecture presented in the paper consists of embedding, encoding, interaction, feature extraction and output layers. The embedding layer produces one vector per word that consists of a pretrained word embedding (which is fine tuned during the training), character-level representation of the word (by applying convolutional layers on character embeddings) and word-level features (part-of-speech tags, and one more "exact match" flag). Encoding layer applies some variant of self-attention on the words in the sentence. Embedding and encoding layers are applied to hypothesis and premise sentences independently. The outputs of encoding layers are merged into a tensor which is passed to a feature extractor. The authors use DenseNet \citep{densenet} architecture as a feature extractor. Its output is passed to a dense layer with softmax activation. 

We took an implementation of the DenseNet architecture from a publicly available repository\footnote{\url{https://github.com/titu1994/DenseNet}} and modified it to fit the current paper. The first difference is that DIIN does not use batch normalization in the DenseNet, as the authors claim that batch normalization slows down the training without improving the accuracy. Another difference is the use of max-pooling layers instead of average-pooling.

It was a little tricky to add an exponential L2 regularization that depends on time, as the regularizers in Keras do not have access to the number of steps during the training. As a workaround we have included L2 penalty directly inside our custom optimizer.

\section{Missing details and inconsistencies}
Few details of the model are not explicitly described in the paper:
\begin{enumerate}
\item The way weights of the network are initialized (this is a very common practice in recent papers).
\item Maximum number of characters in a word.
\item The sizes of character embeddings and convolution kernels applied on them.
\end{enumerate}

We have asked the authors to clarify these points in OpenReview and got the answers for all our questions. Additionally, we found the original implementation in TensorFlow \footnote{\url{https://github.com/YichenGong/Densely-Interactive-Inference-Network}}. Before getting the answers we tried several hyperparameter values. The answers to our questions were the following:

\begin{enumerate}
\item The authors did not explicitly specify initializers, so every layer used TensorFlow's default initializer.
\item The number of characters in a word was limited to 16.
\item Character embedding size was 8. Kernel size of the convolutional layer applied on the embedding was 5 (we guessed this one correctly). Number of filters of convolution was 100. These details were added in the latest revision of the paper.
\end{enumerate}


We trained another model after putting the right hyperparameters.

Then we noticed that the number of parameters in our implementation was significantly different from the number of parameters of the original implementation. To understand why this happens we started to look at the code and discovered few more differences.

\begin{enumerate}
\item There were several differences in the DenseNet implementation:
    \begin{enumerate}
    \item DIIN applies activation function on the output of convolution, while we were applying relu on the input of conv.
    \item DIIN applies decaying L2 regularization on all weights of the network. We applied additional fixed L2 regularization on the kernels of DenseNet convolutions.
    \item DIIN has bias terms in convolutions, while we were ignoring bias terms.
    \item After the last dense-block the DIIN applies max pooling, while we were flattening the output of the last convolution.
    \item The authors write ``Then the generated feature map is feed into three sets of Dense block (Huang et al., 2016) and transition block pair.'' We originally (incorrectly) understood that there were only two transition blocks and three dense blocks. In fact there should have been 3 dense blocks and 3 transition layers. This was the main source of the difference of the number of parameters.
	\item Transition layers do not apply ReLU after convolution, while the paper states that "A ReLU activation function is applied after all convolution unless otherwise noted".
    \end{enumerate}
    
\item We had two mistakes on our side:
	\begin{enumerate}
    \item We set L2 regularization ratio to be $9 \cdot 10^{-6}$ instead of $9 \cdot 10^{-5}$.
    \item We used $0.1$ learning rate instead of $0.5$.
    \end{enumerate}
\end{enumerate}

We have fixed these differences and trained the model one more time. 

After more exploration of the code we found three more differences. 
\begin{enumerate}
\item The paper states that dropout layers are applied before all linear layers and after word embedding layer. However we found that there are several more dropout layers in the code. One of them is applied on the output of interaction (input of DenseNet). Another one is applied on top of character embeddings.
\item The paper states that the optimizer needs to be changed from Adadelta to SGD when the performance does not improve for several steps. The code on GitHub does not switch to SGD (the relevant code is commented out).
\item In order to do model selection, we were evaluating the model on the development set after every epoch (which contains 7849 iterations), while the original implementation is evaluating every 1000 iterations at the beginning and does it more often when the performance is getting higher. The code explicitly checks the performance on the development set and changes the evaluation frequency. This process is not described in the paper.
\end{enumerate}

We did not change the model selection procedure and kept switching to SGD in the late stages of training (to stay consistent with the paper). We tried another run with the additional dropout layers.

\section{Results}
We have trained the network on Stanford's Natural Language Inference dataset \citep{snli:emnlp2015}. The paper also reports results for MultiNLI \citep{MultiNLI} and Quora Question Pairs datasets, but we didn't have time to run the models on those. We also didn't try ensembling.

We have also tried to run the original DIIN code that was available on GitHub. We had a problem testing the final model on the test set (model checkpoints were not saved for some reason), but the performance on the development set was very close to the reported values. 

Table \ref{tab:SNLI} presents the results of our experiments. First two rows are copied from the paper. The third row shows the best results we got by using the original code released by the authors. The remaining rows show the results of our experiments. 

Before we got the exact values for the hyperparameters, we tried several random combinations. We got better results with larger embedding sizes for characters. The best model we could find had character embedding size set to 55, the number of convolutional filters on character-level embeddings was 77, and the model reached 84.48\% accuracy on test set. 

After setting correct hyperparameters values the accuracy of our model dropped to 82.21\%. After updating the DenseNet implementation and fixing the incorrect learning rate the accuracy jumped by four percents. Adding additional dropout layers did not change the accuracy.

\begin{table}[h]
\centering
\begin{tabularx}{\textwidth}{l|r}
 & Test Accuracy \\\hline
DIIN from the paper & \textbf{88.0}\%  \\
DIIN from the paper (ensemble) & \textbf{88.9}\%  \\ \hline
Our rerun of the original code (on dev set) & 88.28\%  \\ \hline
Before setting the correct hyperparameters (our random search) & 84.48\%\\
After setting the correct hyperparameters (based on the authors' answers) & 82.21\%\\
After fixing the learning rate and DenseNet (dev: 85.83\%) & \textbf{86.38}\%\\ 
After adding two more dropout layers & 86.31\%\\\hline
Our custom optimizer (ADAM, then Adadelta, then SGD) & 87.27\%\\\hline
\end{tabularx}
\caption{\label{tab:SNLI}Results on SNLI dataset.}
\end{table}


\begin{figure}[b]
\begin{center}
\includegraphics[width=0.7\textwidth]{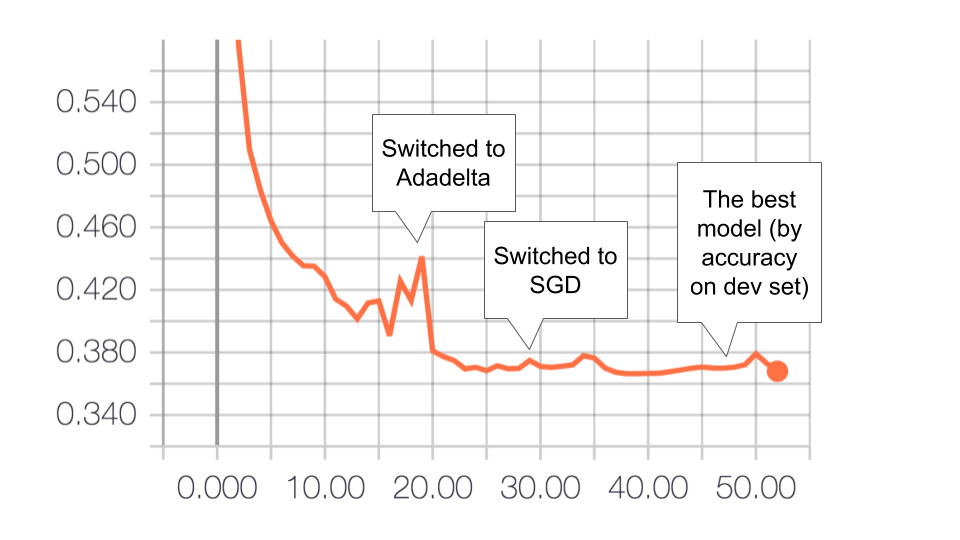}
\caption{Loss on the development set for our best model (it uses three optimizers unlike the original paper)}
\end{center}

\label{fig:val_loss}
\end{figure}

The maximum difference between accuracy numbers on the development and test sets was around 0.5\%.

Finally we played a little with the optimizer and noticed that it has an important role. We obtained our best result with three different optimizers. We started to evaluate the model after every 500 iterations. We started with Adam optimizer\cite{}, switched to Adadelta if loss on the development set did not improve for 3 consecutive epochs, and switched to SGD if the loss did not improve for 4 consecutive epochs (Fig. \ref{fig:val_loss}). This way we got 87.27\% accuracy on the test set after 47 epochs of training. 

\section{Conclusion}
In general, this paper includes most of the important details for reproducing the results. Our models did not reach the accuracy numbers reported in the paper. The reasons could include the way model selection is performed or other differences that we could not find. 

In general, we identify four issues with the paper in terms of reproducibility that are common to many recent papers in deep learning.

\begin{enumerate}
\item First is the reliance on default values of the chosen deep learning framework. For example, most of the time researchers use the default weight initialization strategy provided by the framework. On one hand these default values are not universal among different frameworks, on the other hand the defaults can change in the future versions\footnote{One example is discussed in TensorFlow repository on GitHub: \url{https://github.com/tensorflow/tensorflow/issues/12973}}. We believe it is important to briefly mention the default values and even the exact version of the framework.

\item Many reproducibility issues arise when the authors use other neural architectures as components in their networks and do not describe them in sufficient detail. In this particular case the authored wrote their own implementation of DenseNet in Tensorflow, while we used another implementation of DenseNet (in Keras). Both were different from the official implementations (they are in Torch\footnote{\url{https://github.com/liuzhuang13/DenseNet}} and Caffe\footnote{\url{https://github.com/liuzhuang13/DenseNetCaffe}}). 

\item Many papers do not explicitly describe how the model selection is performed. DIIN paper explicitly mentioned that the main metric is the accuracy on the development set, but didn't mention how often the accuracy was computed (and that the frequency was increased towards the end of the training).

\item One easy trick that the authors can do to ease reproducibility (in case the code is not released) is to mention the exact number of parameters in the network (or even in the individual components). Those numbers can act as good "checksums". There have been some papers in the past that were reporting this number, but it seems to happen less often recently.
\end{enumerate}

Finally, we want to mention that the double-blind strategy of ICLR 2018 (including the possibility of having arxiv preprints) added confusion for the authors about releasing their code. The code was on GitHub (for the arxiv version) but was not cited in the paper as it would violate anonymity.

\bibliography{sample}
\bibliographystyle{iclr2018_workshop}

\end{document}